\documentclass[11pt,a4paper]{article}
\usepackage[hyperref]{acl2017}
\usepackage{times}
\usepackage{latexsym}
\usepackage{booktabs}
\usepackage{amsmath}
\usepackage{float}
\usepackage{amssymb}
\usepackage{graphicx}
\usepackage{tabularx}
\newcolumntype{b}{>{\hsize=1.3\hsize}X}
\newcolumntype{s}{>{\hsize=.7\hsize}X}

\usepackage{url}
\usepackage[section]{placeins}

\aclfinalcopy 
\def\aclpaperid{121} 

\title{Detecting and Explaining Crisis}

\author{Rohan Kshirsagar\\
		Columbia University \\
	    {\tt rmk2161@columbia.edu}  \\ \And 
  Robert Morris \\
  Koko \\
  {\tt rob@itskoko.com} \\ \And
 Samuel R. Bowman \\
 Center for Data Science\\
 and Department of Linguistics\\
 New York University\\
  {\tt bowman@nyu.edu}\\}

\date{}

\begin{document}

\maketitle

\begin{abstract}
Individuals on social media may reveal themselves to be in various states of crisis (e.g. suicide, self-harm, abuse, or eating disorders). Detecting crisis from social media text automatically and accurately can have profound consequences.

However, detecting a general state of crisis without explaining why has limited applications. An explanation in this context is a coherent, concise subset of the text that rationalizes the crisis detection. We explore several methods to detect and explain crisis using a combination of neural and non-neural techniques. We evaluate these techniques on a unique data set obtained from Koko, an anonymous emotional support network available through various messaging applications. We annotate a small subset of the samples labeled with crisis with corresponding explanations. Our best technique significantly outperforms the baseline for detection and explanation. 
\end{abstract}
\section{Introduction}
Approximately one person dies by suicide every 40 seconds \citep{who2016}. It accounts for approximately 1.5 \% of all deaths, and is the second leading cause of death among young adults \citep{who2016}. There are indications that for each adult who dies of suicide there may have been more than 20 others attempting suicide \citep{who2016}. Closely tied to suicide are self-harm, eating disorders, and physical abuse. 13 to 23\% of adolescents engage in self-injury at some point \citep{jacobson2007}. In the United States, about 7 million females and 1 million males suffer from eating disorders annually \citep{eatingdisorder2013} and an average of 20 people are physically abused by intimate partners every minute \citep{abuse2015}. Self-harm victims are more likely to die by suicide by an order of magnitude \cite{selfharm2014}. Additionally, eating disorders and physical abuse increase the risk of suicide \cite{eatingdisorder2013,abuse2015}.

We identify all of these phenomena (suicide, self-harm, eating disorders, physical abuse), with the term crisis. Someone who is in crisis is likely in need of some form of immediate support, be it  intervention, therapy, or emergency. Roughly a third of people who think about suicide make a plan; 72\% of those who report making a suicide plan actually make an attempt \cite{kessler1999}. 

Accurate, automatic detection of someone in crisis in social media, messaging applications, and voice assistants has profound consequences. A crisis classifier can enable positive outcomes by enabling human outreach at earlier stages, and rescues at later stages. 

In many ways, however, it can still fall short of a human detector, by way of lacking an explanation or rationale of why classifier detected crisis. The factors that explain why someone is in crisis can range from suicidal ideation to eating disorders, from self-harm to sexual abuse.

In crisis situations, triage can improve if the detection system can explain why the person is crisis.
Someone who is about to die by suicide via overdose should receive a different response than someone who is considering anorexia 

due to self-image issues. Population level surveillance, diagnostics, and statistics are much improved due to factor based explanation. Finally, in human-in-the-loop crisis systems, the human responder can better sift through information if the factors of crisis were visually highlighted through automated means \cite{dinakar2014}. With the rise of complex neural models in classification tasks, 
we've seen gains in accuracy at the cost of transparency and interpretability \citep{kuhn2013}. 
In this work, we present models that we validate both for their raw accuracy and for the quality of their explanations.
Validating a model's explanations, in addition to its detection performance, can improve interpretability in the model. In summation, automatically generating explanations for crisis detection scales and pays off in many ways. 

While we evaluate our models' explanations against human reference explanations, it is not practical to collect enough explanations to train these models on such data. Collecting an explanation requires an annotator to write free text or to highlight text for every case of crisis, sometimes more than once for a post (e.g. for a sexual abuse victim considering suicide), while merely identifying crisis is a simple binary decision task that can be performed much more quickly and cheaply. 

In this paper, we explore the problem of generating explanations for crisis without explicit supervision using modern representation learning techniques. We demonstrate our success comparing our proposed models with a variety of explanatory methods and models on a rationale-labeled test set. We evaluate the generated explanations through ROUGE metrics against human-produced references. In addition, we show detection performance that outperforms prior methods. 

\section{Related Work}

\paragraph{Detecting Crisis}
\citet{wood2016} identify 125 Twitter users who publicly state their suicide attempt online on a specific date and have tweets preceding the suicide attempt. They artificially balance these tweets using data from random users who are assumed to be neurotypical, acknowledging that this data will be contaminated with users who also ideate and attempt suicide. They train simple, linear classifiers that show promise in detecting these suicidal users and discuss the difficulties of realizing this technology, highlighting privacy and intervention concerns. In our work, we attempt detection and explanation on phenomena that includes but is not limited to suicide on a dataset that is significantly larger and not artificially balanced. However, we do not incorporate the record of suicide attempt as signal when labeling. 

\citet{gkotsis2016} operate on a filtered subset of mental health records to determine whether a mention of suicide is affirmed or negated. They do classification on mental health records, 
which are filtered by the \texttt{suicid*} keyword. The goal of their work was the development of improved information retrieval systems for clinicians and researchers, with a specific focus on suicide risk assessment. Thus, the domain is constrained. Their work also differs from ours significantly in its technical execution. Rather than use neural network classifiers, they use probabilistic context free grammars to execute negation detection. This task is quite different than ours, both in dataset and approach, and is most likely not applicable to open-domain social media text. They also do not aim to or need to provide explainable detections, as mentions of suicide are clearly present in all of their data and negation detection is sufficient 'rationale' for affirming or negating that mention. 

\citet{tong2014} annotate Twitter data for suicide-risk with the level of distress as the label and achieve high inter-annotator agreement. 
They use a combination of specialized keyword search and LIWC sadness scores \citep{pennebaker2001} 
to filter 2.5 million tweets down to 2000 in order to make the annotation task tractable. Our source dataset, which we introduce in the next section, has a significantly higher base rate of crisis; thus, no filtering is necessary. 
They train SVM classifiers on bag of n-grams to detect distress on different subsets of annotations, but do not explore neural classifiers, nor unsupervised explanations of detections. 

\citet{lehrman2012} and \citet{odea2015} also detect distress on small datasets using simple classifiers. \citet{lehrman2012}  annotate 200 samples for distress level and discretize counts related to bag of word, part of speech, sentence complexity and sentiment word features to train a variety of multiclass classifiers. \citet{odea2015} annotated nearly 2000 tweets for different levels of suicidality and used word counts as features, filtered by document frequency.  
In our work, we compare neural techniques against linear models trained on word frequency counts both for detection and explanation as a baseline. Due to the relatively large amount of data in our training set, we do not use any custom features for the baseline. 

\citet{mowery2016} detect depression in Twitter data in two stages: 1) detecting evidence of depression at all and 2) classify the specific depressive symptom if depression was detected. 
This is a kind of explanation in that it directly detects one of three symptoms of depression (fatigue, disturbed sleep, depressed mood). However, their data is explicitly annotated for these sub-factors, whereas our data is not. 1,656 tweets in their dataset were annotated with specific depressive symptoms. 

\paragraph{Interpretable Neural Networks}
In the past few years, neural attention mechanisms over words \cite{bahdanau2014} have led to improvements in performance and interpretability in a range of tasks, such as translation (Bahdanau) and natural language inference \citep{rocktaschel2015}. These models induce a soft alignment between two sequences with the primary aim of using it to remove an information bottleneck, but this alignment can be also be used quite effectively to visualize which inputs drive model behavior.

\citet{lei2016} present a more direct method for training interpretable text classifiers. Their model is also trained end-to-end, but instead of inducing a soft weighting, it extracts a set of short subspans of each input text that are meant to serve as sufficient evidence for the final model decision. 
In another work with similar goals, \citet{ribeiro2016} introduce a model agnostic framework for intepretability, LIME, that learns an interpretable model over a given sample input that is locally faithful to the original trained model. 

\section{Methods}
Our training set consists of $N$ examples $\{X^i, Y^i\}_{i=1}^N$ where the input $X^i$ is a sequence of tokens $w_1, w_2, ..., w_T$, and the output $Y^i$ is a binary indicator of crisis.

\subsection{Word Embeddings}
 Each token in the input is mapped to an embedding. We use reference GloVe embeddings trained on Twitter data \cite{pennington2014}. We used the 200 dimensional embeddings for all our experiments, so each word $w_t$ is mapped to $x_t \in {\mathbb R}^{200}$. We denote the full embedded sequence as $x_{1:T}$. 
\subsection{Recurrent Neural Networks}
A recurrent neural network (RNN) extends on a traditional neural network to recursively encode a sequence of vectors, $x_{1:T}$, into a sequence of hidden states. The hidden state of the RNN at $t-1$ is fed back into the RNN for the next time step.
\begin{align*}
h_t = f(x_t, h_{t-1};\Theta)
\end{align*}
This allows the network to construct a representation incrementally as it reads the input sequence. In particular, we encode the sequence using a gated recurrent unit \citep[GRU;][]{cho2014} RNN. The GRU employs an update gate $z_t$ and reset gate $r_t$ that are used to compute the next hidden state $h_t$ 
\begin{align*} 
h_t = (1 - z_t)h_{t-1}  + z_t \tilde{h}_t\\
z_t = \sigma(W_z\textbf{x}_t + U_z\textbf{h}_{t-1}) \\
\tilde{h}_t = tanh(W\textbf{x}_t + U(\textbf{r}_t \odot \textbf{h}_{t-1}))\\
r_t = \sigma(W_r\textbf{x}_t + U_r\textbf{h}_{t-1})
\end{align*} 
We use a bidirectional RNN (running one model in each direction)  and concatenate the hidden states of each model for each word to obtain a contextual word representation $h^{bi}_t$. 

\subsection{Attention over Words}
With attention, a scoring function scores the relevance of each contextual word representation $h^{bi}_t$. We employ the unconditional attention mechanism used to do document classification employed by \citet{yang2016}.

\begin{align*} 
u_t = tanh(W_wh^{bi}_t + b_w)\\
\alpha_t = \frac{\exp{u_t}}{\sum\limits_{t} \exp{u_t}} \\
d = \sum\limits_{t} \alpha_t h_t
\end{align*}

The attention mechanism serves two purposes. $d$ acts as a contextual document representation which can be fed into a downstream model component for detection. In addition, the score vector  $u_{1:N}$, can be utilized to seed our explanation, which will be expanded on in a following section. Optionally for detection, we encode the document by using the last hidden state of a single forward GRU, without the reverse GRU and attention mechanism. Both encoding schemes are evaluated in our experiments.

\subsection{Training Objective}

The final document encoding of each sample, $d$, is fed into a sigmoid layer with one node to detect the probability of crisis. 
We minimize the logistic loss objective during training. 
\begin{align*}
l(y,p) = -ylog(p) - (1 - y)log(1-p) \\
\end{align*}
where $y$ is the true value and $p$ is the output of the logistic output layer. 

\subsection{Seeding the Explanations}
Our next goal is to generate explanations given the inputs and outputs for our trained model. We do this by building a subset of words which `seed' the explanation generation function. The explanation generation function is fixed while testing across all seeding techniques, thus allowing extensibility by modularizing the seed function using a relevant model. The seed function is meant to give a set of tokens from the input that most influenced the prediction, thus sewing the initial stitches of the explanation. For the task of detecting crisis, descriptive content words, such as adjectives, nouns, and verbs, are desirable compared to stop words or punctuation. 

We test three techniques of seeding words for a given input: (1) Using the magnitudes of activated coefficients in a logistic regression model. This acts as our baseline. (2) Using the distribution of attention from our neural model. (3) Using LIME, which can generate words that led to a prediction for any model. Each seed function is passed in the number of seed words to return, $k$. This allows us to maintain similar output behavior for all three techniques; it also allows us to extend the seed functions to more complex models. 
We will now detail how seeding works for each of these mechanisms.

\begin{enumerate}
\item \textbf{Logistic Coefficients:} Logistic regression is a linear model that learns a vector of weights for a fixed set of features to detect in binary classification. As a baseline, we train a logistic regression model on unigrams to learn a vector of weights for each word in the vocabulary. For our seed function, we find the $k$ most highly-weighted activated features according to the model. A feature is activated if the word occurs in the given input. 
\item \textbf{Neural Attention:} In this setting, we select seeds by sorting the words by their attention weights $u$. In order to get human interpretable scores for attention, we introduced a configurable dial to control how attention was distributed over the input by introducing an L2 penalty on the output of the attention. 
\item \textbf{LIME:} The LIME API contains a \texttt{num\_features} parameter in the \texttt{explain\_instance} function. Each explanation will then result in learning an interpretable model, which can be used to then seed the explanation. 
The LIME API is applied to both models, the baseline logistic and the neural model. 
\end{enumerate}

\subsection{Explanation Generation Algorithm}

We use a novel algorithm for producing explanations that depends on seeds from a separately-developed seeding module. 
The algorithm acts on the input text and the $k$ explanation seeds. It works as follows. First, the sentence of importance is identified by taking the sentence with the most seeds. The identified sentence is then parsed with a dependency parser \cite{honnibal-johnson:2015:EMNLP}, and traversed from the root to find the highest seed in the sentence.  If the highest seed token is not a verb and not the head of the entire sentence, we then traverse to the seed's head node. Subsequently, the subtree phrase of the highest seed is used for the explanation. Since the parse is projective, the subtree is necessarily a contiguous sequence of tokens.

\section{Experiments}
\subsection{Training Data}

\begin{figure*}
  \includegraphics[width=\textwidth,height=2cm]{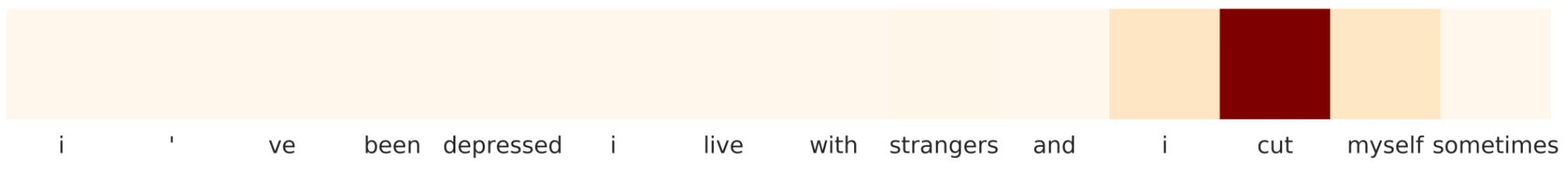}
  \includegraphics[width=\textwidth,height=3cm]{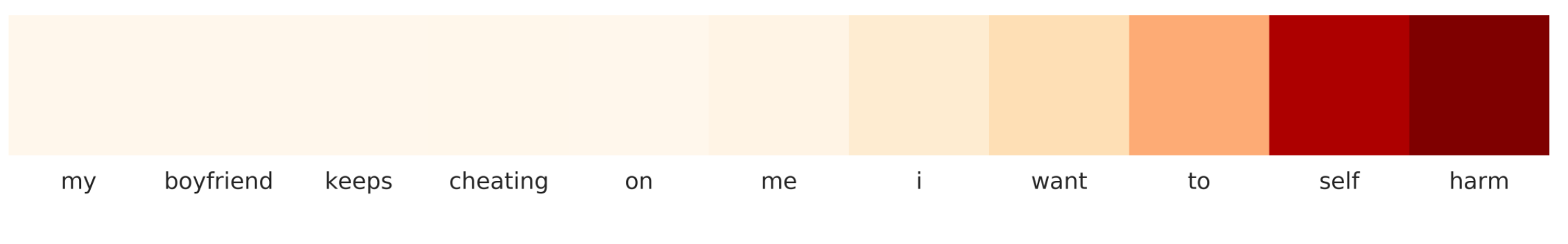}
  \caption{Visualizing Attention for crisis.}
\end{figure*}

Koko has an anonymous peer-to-peer therapy network based on an clinical trial at MIT \cite{morris2015}, which is made available through chatbots on a variety of social media platforms including Facebook, Kik, and Tumblr. They provided us with our training data through a research partnership. The posts on the platform generally come from users who are experiencing negative thoughts and need some form of emotional support. Each post is on average 3.1 sentences long with a standard deviation of 1.7 sentences. The training set is roughly 106,000 binary labeled posts (crisis or not).

Their data was annotated for crisis by crowdworkers. 
During annotation, annotations were given clear instructions on what consists of crisis, examples, and common mistakes and helpful tips. These instructions were revised over multiple iterations of small batches of data to improve inter-annotator agreement. Using a minimum of three labelers per sample, they achieve over 95\% inter-annotator agreement.

Because the platform is a support network, the rates of depression and other mental disorders are high: the annotation task identified roughly 20\% of the training data as crisis. This is in contrast to previous work using Twitter data, where multiple layers of filtering are required to get a reasonable sample of distress \cite{tong2014}. Our dataset requires no filtering and estimates the natural distribution of the platform. 
\subsection{Explanation data}
We have select a set of 1242 labeled posts as our test set. Of these, 200 are labeled crisis. 
We annotate the 200 crisis samples with their corresponding explanations. An explanation is a phrase or clause in the post that most strongly identifies the rationale behind the crisis label. When selecting the explanation, we aim for them to be accurate, coherent, and concise. 

\subsection{Model Configuration and Training}
We tokenize the data using Spacy \citep{honnibal-johnson:2015:EMNLP}. We do not fine-tune the pretrained GloVe embeddings, but rather learn a simple embedding transformation matrix that intervenes between the embeddings and the RNNs. We use 200 dimensional embeddings and 100 dimensional forward and backward GRUs (yielding 200 dimensional contextual representations). We apply an L2 penalty on the attention output using $\lambda = .0001$. We pad each input to 150 words. We train using RMSprop with a learning rate of .001 and a batch size of 256. We add dropout with a drop rate of 0.1 in the final layer before detecting to reduce overfitting. We determined the dropout rate, batch size, and input length empirically through random hyperparameter search and determined $\lambda$ for the attention penalty using human evaluation.
We use the best model from 20 epochs of training selected using a validation sample of 10\% of the source data (excluding the test data).

\section{Results and Discussion}

\begin {table}[t]
\centering
\begin{tabular}{lrrr}
\toprule
{} &         Precision &         Recall  &         F1 \\
\midrule
logistic      &  \textbf{0.87}  & 0.53  & 0.66 \\
rnn+attention &  0.85  & 0.69 & 0.76 \\
best rnn      &  0.82 & \textbf{0.77} & \textbf{.80} \\
\bottomrule
\end{tabular}
\caption {Crisis Detection Performance on Test Data 
} \label{tab:title} 
\end {table}
\subsection{Detection Evaluation}
The neural models significantly outperform the logistic model in detection accuracy (Table 1), with the best neural model achieving a .80 F1 on the crisis detections, compared to .66 for the logistic model. The neural attention model achieves a .76 F1 score, which is still significantly better than the linear baseline. The best model does not have an attention penultimate layer, bur rather a single feedforward GRU layer. 

\subsection{Attention Visualization}
We first validate that the attention mechanism yields distributions that meet our expectations. This is done by visualizing the attention using a heat map, with each normalized attention weight aligned with the corresponding token in the input. Initially, we found that the attention distribution had a very low entropy, placing the bulk of the probability in a single token of the input. We penalized low entropy outputs using an L2 penalty, controlled by a $\lambda$ parameter.
 We did not further tune it to boost explanation evaluation scores, though we expect this could improve performance. Figure 1 demonstrates the attention output for two crisis samples. For the first sample, we see that attention is focused around the final clause, and is not concentrated entirely on one word. As one would expect, ``i cut myself'' fetches the highest weight in the attention distribution. The second visualization shows singular attention on the word `suicide', thus placing markedly less importance on the rest of the input. This differentiation between background information and crisis signal provides a reassuring signal that the model is using reasonable features. 

\begin {table}\centering
\begin{tabularx}{\linewidth}{sb}
\toprule
\textbf{Text}  & Im really lonely and i want someone who loves me cares for me and i love (ima guy) i want to kill myself because i cant get a girlfriend  \\
\textbf{Gold}  &  kill myself\\ 
\textbf{logistic+coef}&  to kill myself  \\
\textbf{logistic+LIME }   & to kill myself  \\
\textbf{neural+attention} & to kill myself \\
\textbf{neural+LIME }& to kill myself\\
\midrule
\textbf{Text}  & I have to face many changes in the next few months but I'm not ready. Instead I hide in fast food and tv shows. I'm scared that my depression will come back and turn to suicidal thoughts.. Big changes ahead make me worried about suicidal thoughts overwhelming me so I hide.   \\
\textbf{Gold}  & suicidal thoughts\\
\textbf{logistic+coef}& Big changes ahead  \\
\textbf{logistic+LIME }   & I have to face many changes in the next few months but I'm not ready. \\
\textbf{neural+attention} & about suicidal thoughts overwhelming me so I hide \\
\textbf{neural+LIME }& suicidal thoughts\\
\midrule
\textbf{Text } & My parents want me to be a perfect child but I have depression and anxiety. Suicide    \\
\textbf{Gold } &  Suicide\\
\textbf{logistic+coef}&  me to be a perfect child  \\
\textbf{logistic+LIME}    &  me to be a perfect child  \\
\textbf{neural+attention} & I have depression and anxiety. Suicide \\
\textbf{neural+LIME} & I have depression and anxiety.\\
\midrule
\textbf{Text}  & Everyone at school is calling me a nerd,bitch,loser etc... The problem is that I'm starting to believe them and lately I've started cutting..I'm gonna go insane or lose myself   \\
\textbf{Gold } &  I've started cutting\\
\textbf{logistic+coef}& Everyone at school is calling me a nerd, bitch, loser etc..	 \\
\textbf{logistic+LIME  }  & cutting  \\
\textbf{neural+attention} &lately I've started cutting. \\
\textbf{neural+LIME }& lately I've started cutting.\\
\bottomrule
\end{tabularx}

\caption {Explanation Samples} \label{tab:title} 
\end {table}
\subsection{Qualitative Explanation Results}

Interestingly, all of the techniques resulted in several high quality explanations. We surveyed about 20 samples and for each one, at least one of the seeding functions contained the correct explanation. Surprisingly, the logistic baseline performed quite well in this capacity. In Table 2, 
we show an example where all of the techniques got the identical result. This is likely due to the predictive power of the phrase `kill myself'. In many cases, the generated explanation contained more text than is necessary to accurately capture the gold explanation. The second example (Table 2) shows this in the neural+attention technique. This may suggest room for improvement in the explanation generation technique. The third example
shows a difficult case in which the majority of the text is background information and only the last word of the input is included in the gold explanation. 
We see that both neural models and logistic+LIME are successful in capturing roughly the correct explanation.

\subsection{Quantitative Explanation Results}

We evaluate the generated explanations using ROUGE-1 and ROUGE-2 \citep{lin2004rouge}, which measure the overlapping units (unigrams and bigrams respectively) of the generated text and reference texts. In Table 3 and 4, the average ROUGE-1 and ROUGE-2 scores for the generated explanations are listed for each model and seed strategy.  By a large margin, the neural classifier\footnote{We use the RNN with attention in this result. The forward RNN  in conjunction with LIME showed nearly identical ROUGE performance.} in conjunction with the LIME seed function outperformed the rest of the models. In ROUGE-2 evaluation, it beats the next best average F1 score by a margin of 10 points and in ROUGE-1 evaluation, it beats the next best average F1 by 12 points. Since LIME directly determines which input most influences the prediction, while attention does so only  indirectly, this result makes sense. However, the LIME seeding function is the slowest approach we consider, taking up to a minute to generate a explanation. The neural attention seeding is negligible in contrast to this. 
In Table 3, the ROUGE Metrics show similar performance for the baseline logistic model and the neural model. However, in Table 1, we see that detection output is much better for the neural models. This suggests that though the logistic regression is quite reasonable in ranking features by weights, it fails to capture subtleties and dependencies in a sequence that an RNN captures. Thus, neural+attention 
is a better choice between the two. The logistic+LIME outperforms the baseline by 5 points in precision for ROUGE-1 and around 3.5 points in precision for ROUGE-2. This exemplifies the efficacy of LIME, which is tuned for the individual example, rather than the model coefficients, which are tuned for the training data. 

\begin {table}[t]
\tabcolsep=0.11cm
\begin{tabular}{lrrr}
\toprule
{}  &         Precision &         Recall &         F1 \\
\midrule
logistic+coef   &  0.358 &  0.590  &  0.396  \\
logistic+LIME  &  0.409 &  0.610 &  0.432\\
neural+attention &  0.360 &  0.605 &  0.406\\
neural+LIME     &  \textbf{0.492} &  \textbf{0.745} &  \textbf{0.536} \\
\bottomrule
\end{tabular}
\caption {ROUGE-1 Scores} \label{tab:title} 
\end {table}

\begin {table}[t]
\tabcolsep=0.11cm
\begin{tabular}{lrrr}
\toprule
{} &             Precision &         Recall&  F1   \\
\midrule
logistic+coef     &  0.267 &  0.475 &  0.289 \\
logistic+LIME  &  0.301 &  0.478 &  0.311\\
neural+attention&  0.286 &  0.485 &  0.309  \\
neural+LIME      & \textbf{ 0.397} &  \textbf{0.615}  & \textbf{ 0.413}\\
\bottomrule
\end{tabular}
\caption {ROUGE-2 Scores} \label{tab:title} 
\end {table}

\section{Conclusion}

In this paper, we present and compare explanation-oriented methods for the detection of crisis in social media text. We introduce a modular approach to generating explanations and make use of neural techniques that significantly outperform our baseline. The best models presented are both effective at detection and produce explanations similar to those produced by human annotators. We find this exciting for two reasons: 
Within the domain of crisis identification, successes in explanation help to build the trust in trained models that is necessary to deploy them in such a sensitive context. 
 Looking beyond this, we expect that our techniques may generalize to text classification more broadly. In the future experiments, we hope to explore  human evaluation of the generated explanations as an indicator of trust in the model, to investigate compression-based approaches to explanation \citep{lei2016}, and to consider richer architectures for text classification.

\section*{Acknowledgments}
We thank the anonymous reviewers and Kareem Kouddous for their feedback. Bowman acknowledges support from a Google Faculty Research Award and gifts from Tencent Holdings and NVIDIA Corporation. We thank Koko for contributing a unique dataset for this research.

\bibliography{naaclhlt2016}
\bibliographystyle{acl_natbib}

\end{document}